\documentclass[letterpaper]{article} 
\usepackage{aaai2027}  
\usepackage[hyphens]{url}  
\usepackage{graphicx} 
\usepackage{natbib}  
\usepackage{caption} 
\usepackage{algorithm}
\usepackage{algorithmic}
\usepackage{amsmath}

\usepackage{newfloat}
\usepackage{listings}
\DeclareCaptionStyle{ruled}{labelfont=normalfont,labelsep=colon,strut=off} 
\floatstyle{ruled}
\newfloat{listing}{tb}{lst}{}
\floatname{listing}{Listing}

\usepackage{booktabs}
\usepackage{multirow}
\title{InfiHall-Bench: A Benchmark for Hallucination Evaluation in Autoregressive Text-to-Image Generation}
\author{
    Yuxia Qiao,
    Xinhao Zhong,
    Junhao Li,
    Yi Sun,\\
    Jiaxin Hong,
    Shuoyang Sun,
    Bin Chen
}

\affiliations{
    Harbin Institute of Technology, Shenzhen
}

\begin{document}
\maketitle
\section{Abstract}
Hallucination benchmarks for text-to-image generation are predominantly
designed around the capability profile of diffusion models, which limits both
their diagnostic value for the emerging class of visual autoregressive (VAR)
models and their usefulness for developing mitigation methods. We identify two
limitations of current prompt construction paradigms. First, a substantial
portion of existing prompts is no longer difficult enough for modern
generators. Second, category-based construction instantiates prompts through
recurring object--attribute--relation templates and therefore rarely tests how
several visual requirements interact within a single image. We introduce
\textbf{InfiHall-Bench}, a controlled benchmark of 411 prompts obtained by
systematically composing five foundational, explicitly stated, and visually
verifiable skills: attribute binding, spatial relation, action relation, part
consistency, and scene consistency. Each prompt carries 4.05 annotated skills
on average, so the benchmark measures hallucinations under diverse
combinations of visual requirements rather than one requirement at a time.
Using this benchmark, we find that hallucinations in VAR models exhibit a
scale-wise persistence pattern: semantic errors can become established at
different autoregressive scales and are rarely repaired by subsequent
fine-scale steps. Motivated by this observation, we propose
\textbf{Confidence-Masked Hallucination Mitigation
(CMHM)}, which estimates the prediction confidence of target visual tokens
at every autoregressive scale during teacher-forced fine-tuning and
downweights the tokens that are already predicted reliably, so that
optimization concentrates on uncertain predictions wherever they occur. Experiments show that
InfiHall-Bench separates models that existing compositional benchmarks rate
as comparable, and that CMHM consistently improves over uniform LoRA
fine-tuning for VAR-based text-to-image generation.

\section{Introduction}

Text-to-image (T2I) generation has advanced rapidly with large-scale
generative models. Diffusion models improve image quality, diversity, and
text--image alignment through iterative denoising in continuous latent
spaces~\cite{rombach2022high,saharia2022photorealistic,ramesh2022hierarchical,podell2024sdxl}.
More recently, visual autoregressive (VAR) models have emerged as an
alternative paradigm that represents images as discrete visual tokens and
predicts them progressively from coarse to fine scales~\cite{tian2024visual}.
Infinity further introduces bitwise autoregressive modeling for efficient
high-resolution synthesis~\cite{han2025infinity}.

Despite this progress, T2I models often generate visually plausible images
that fail to satisfy explicit prompt requirements. Following prior work on
prompt--image faithfulness~\cite{hu2023tifa,lim2025evaluating,huang2024t2i}, we use the
term \emph{hallucination} for any content in the generated image that
contradicts or omits a requirement stated explicitly in the prompt: a wrong
attribute binding, a missing object or object part, an incorrect spatial
layout, or an unrealized interaction. Existing benchmarks measure such
failures through complementary protocols.
T2I-CompBench~\cite{huang2023t2i} assesses attribute binding, object
relationships, and complex compositions, while GenEval~\cite{ghosh2023geneval}
verifies object presence, count, position, and color.
TIFA~\cite{hu2023tifa} decomposes prompts into question--answer pairs for
interpretable faithfulness evaluation. More recent benchmarks broaden the
scope: GenAI-Bench~\cite{li2024evaluating} covers both basic compositions and
advanced skills such as counting, comparison, negation, and universality,
while T2I-ReasonBench~\cite{sun2026t2i} targets knowledge-intensive generation
involving idiom interpretation, textual image design, entity reasoning, and
scientific reasoning.

\begin{figure*}[t!]
    \centering
    \includegraphics[width=0.9\textwidth]{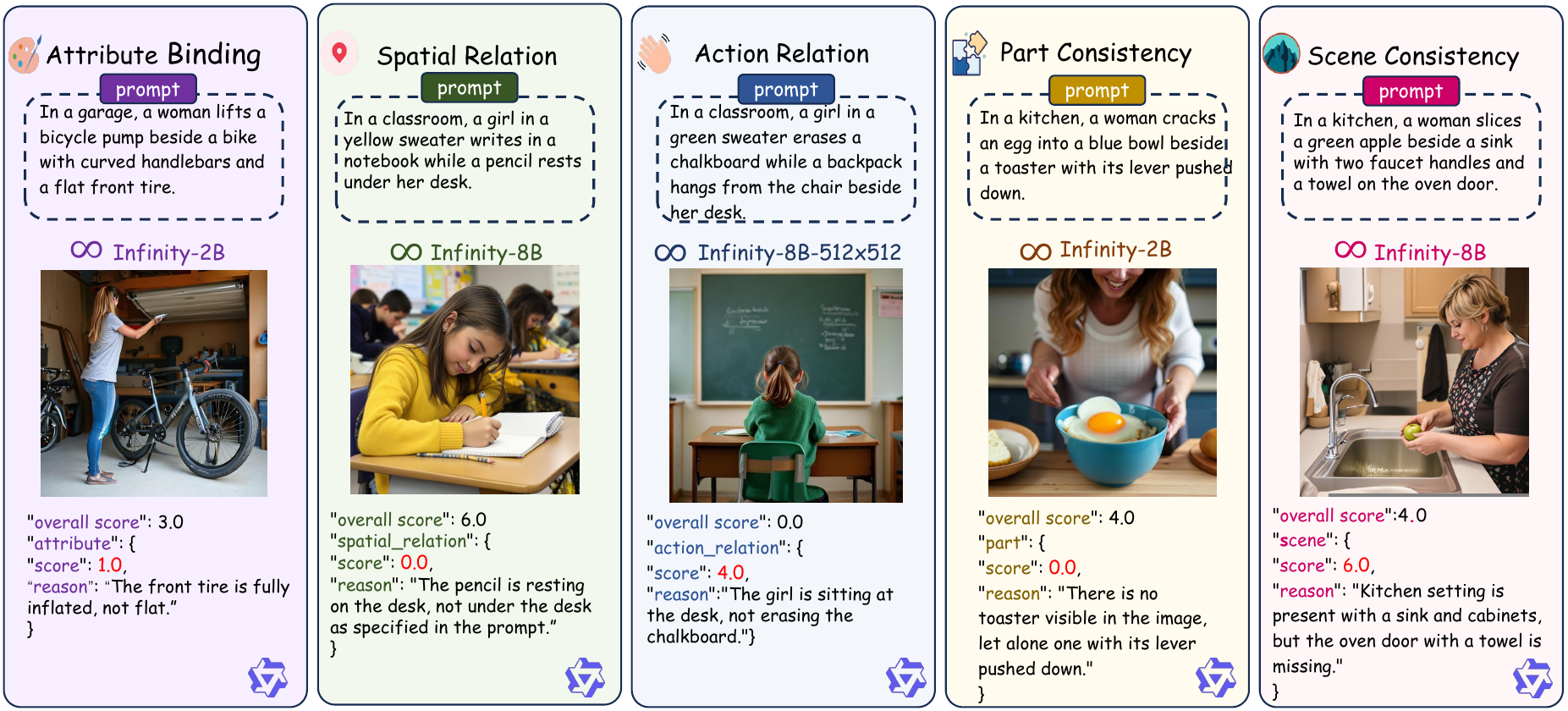}
    \caption{\textbf{Overview of InfiHall-Bench.} Representative failures
    across the five skills covered by the benchmark. Each example shows the
    prompt, an Infinity output, and the skill-specific feedback returned by
    the evaluator.}
    \label{fig1}
\end{figure*}

Although existing benchmarks provide broad taxonomies and valuable evaluation protocols, their limitations extend beyond the granularity or sophistication of the categories they cover. More fundamentally, most were developed around diffusion models and evaluate final-image alignment using categories shaped by diffusion-era capability profiles. Their diagnostic value does not necessarily transfer to visual autoregressive (VAR) models: diffusion models iteratively refine continuous latent representations~\cite{rombach2022high}, whereas VAR models ~\cite{tian2024visual} progressively predict discrete visual tokens from coarse to fine scales, introducing scale-wise decisions that are not exposed by category-level endpoint scores. Consequently, similar aggregate results across diffusion and VAR models, or across VAR variants with different capacities and output resolutions, do not necessarily indicate similar hallucination behavior. A useful benchmark should therefore distinguish increasingly capable models while revealing architecture- and scale-dependent failure profiles. However, many existing benchmarks construct prompts from category-specific templates, resulting in recurring object--attribute--relation patterns~\cite{ghosh2023geneval,huang2023t2i}. Human-authored prompts improve linguistic naturalness and semantic diversity, but do not guarantee systematic coverage of interactions among multiple visual constraints (Li et al. 2024). Strong performance on individual categories may therefore mask failures in jointly satisfying several explicit requirements within a single prompt. Increasing difficulty through implicit semantics, world knowledge, or higher-order reasoning does not resolve this issue, because it confounds failures in visual realization with failures in prompt comprehension or reasoning. These limitations motivate a controlled benchmark that systematically composes diverse foundational constraints while keeping them explicit and directly verifiable from the generated image.

Motivated by this gap, we revisit benchmark construction and study
hallucination behavior in Infinity, the strongest publicly available VAR
generator. Our analysis yields two findings. First, prompt difficulty does not
follow from membership in an advanced category: combinations of explicit
attributes, layouts, actions, object parts, and scene-level constraints
continue to expose substantial failures, which suggests that jointly realizing
multiple visual requirements is itself the hard part. Second, hallucinations exhibit a scale-wise persistence pattern: once a semantic error becomes established at one autoregressive scale, subsequent scales tend to refine rather than correct it. Because the scale at which an
error emerges varies across samples, these findings motivate controlled
multi-skill prompt construction and confidence-guided optimization across all
scales for hallucination mitigation.

We introduce InfiHall-Bench, illustrated in Fig.~\ref{fig1}, for diagnosing hallucinations under explicit visual requirements. It covers five foundational skills: attribute
binding, spatial relation, action relation, part consistency, and scene
consistency. Rather than extending the taxonomy toward more
reasoning-intensive tasks, these skills serve as controlled and visually
verifiable units from which diverse multi-skill prompts are composed. Several
skills jointly guide an LLM to generate each candidate, which is then passed
through automatic filtering, manual inspection, and distributional
rebalancing. Every annotated skill is scored separately, and an overall score
measures holistic prompt satisfaction. The benchmark applies to any T2I model;
we use it primarily to characterize the VAR family, and report diffusion
models as reference points.

With the proposed InfiHall-Bench, we further analyze hallucinated samples and
find that semantic errors can emerge at different visual scales and persist
through subsequent coarse-to-fine refinement. Motivated by
this behavior, we propose Confidence-Masked Hallucination Mitigation
(CMHM), a lightweight LoRA-based~\cite{hu2022lora} approach for mitigating
hallucinations in Infinity. During teacher-forced fine-tuning, CMHM estimates the prediction confidence of target visual tokens at every autoregressive scale and downweights the
tokens that are already predicted reliably. This prevents well-learned
predictions from dominating the objective and focuses adaptation on uncertain
visual tokens wherever they occur, without modifying the generation
architecture.
\begin{figure*}[t!]
    \centering
    \includegraphics[width=0.9\textwidth]{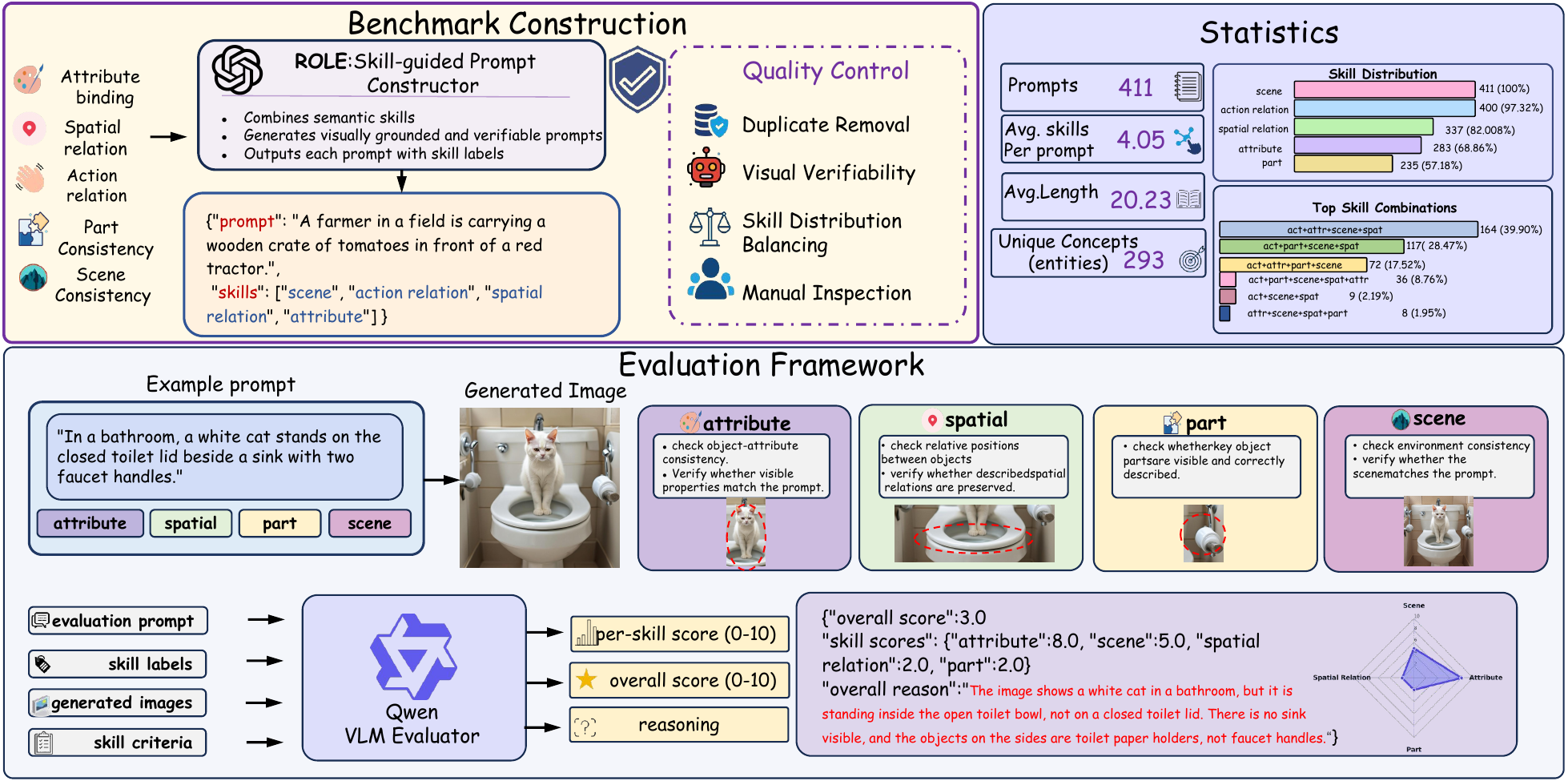}
    \caption{\textbf{Benchmark construction and evaluation.} The upper panel shows skill-guided prompt construction, followed by quality control. The lower panel illustrates the evaluation protocol to produce an overall score, per-skill scores, and reasons.}
    \label{fig2}
\end{figure*}
Our contributions are summarized as follows:
\begin{itemize}
\item We revisit T2I hallucination benchmarks in the context of increasingly
capable VAR models and identify two limitations of their prompt design: many
prompts are no longer challenging, and category-driven construction yields
recurring structures with insufficient coverage of the interactions among
multiple explicit visual requirements.
\item We introduce InfiHall-Bench, a fine-grained hallucination benchmark that
evaluates multi-skill prompt--image consistency through skill-guided prompt
construction and skill-conditioned multimodal evaluation.
\item We propose Confidence-Masked Hallucination Mitigation (CMHM), a
LoRA-based mitigation approach that exploits token-level confidence to focus
optimization on hallucination-prone visual predictions.
\end{itemize}

\section{Related Work}
\label{relatedwork}
\paragraph{Text-to-image generation.}
Text-to-image generation has evolved from GAN-based
approaches~\cite{zhang2016stackgan,xu2018attngan,zhang2018stackgan++} to
diffusion
models~\cite{nichol2021glide,saharia2022photorealistic,rombach2022high,podell2024sdxl}
and autoregressive
models~\cite{ramesh2021zero,esser2021taming,chang2022maskgit}. Diffusion
models generate images through iterative denoising and have substantially
improved image quality and text alignment. Autoregressive models instead
represent images as discrete tokens and predict them sequentially or
progressively. VAR~\cite{tian2024visual} adopts coarse-to-fine next-scale
prediction, while Infinity~\cite{han2025infinity} uses bitwise autoregressive
modeling for high-resolution image synthesis. Our work examines hallucinations
in images generated by Infinity.

\paragraph{Hallucination benchmarks in text-to-image generation.}
Existing benchmarks study prompt--image consistency from complementary
perspectives. T2I-CompBench~\cite{huang2023t2i} evaluates compositional
generation through attribute binding, object relationships, and complex
compositions. GenEval~\cite{ghosh2023geneval} focuses on object presence,
counting, spatial relations, and attribute correctness, while
TIFA~\cite{hu2023tifa} introduces a question-answering framework for
interpretable text--image faithfulness evaluation.
GenAI-Bench~\cite{li2024evaluating} broadens the scope through professionally
authored prompts that cover both basic composition and advanced skills,
including counting, comparison, differentiation, negation, and universality.
DALL-Eval~\cite{cho2023dall} and T2I-ReasonBench~\cite{sun2026t2i} further
investigate reasoning-intensive generation, while
I-HallA~\cite{lim2025evaluating} and
T2I-FactualBench~\cite{huang2024t2i} explicitly study hallucination and
factuality in text-to-image models.
\begin{table*}[t!]
    \centering
    \small
    \setlength{\tabcolsep}{6.0pt}
    \begin{tabular}{l|cccccc}
    \toprule
    \textbf{Model} & Color & Shape & Texture & Spatial & Non-spatial & Complex \\
    \midrule

        \multicolumn{7}{c}{\textbf{Diffusion T2I Models}} \\
        \midrule

        Stable Diffusion v1.4
        & 0.630 & 0.676 & 0.661 & 0.089 & 0.309 & 0.459 \\

        Stable Diffusion v2
        & 0.682 & 0.707 & 0.706 & 0.146 & 0.309 & 0.479 \\

        \midrule
        \multicolumn{7}{c}{\textbf{Visual Autoregressive T2I Models}} \\
        \midrule

        Infinity-2B
        & 0.879 & 0.726 & 0.804 & 0.217 & 0.311 & 0.513 \\

        Infinity-8B-512
        & 0.904 & 0.759 & 0.855 & 0.263 & 0.314 & 0.520 \\

        Infinity-8B
        & \textbf{0.921} & \textbf{0.813} & \textbf{0.892} 
        & \textbf{0.368} & \textbf{0.316} & \textbf{0.524} \\

        \bottomrule
    \end{tabular}

    \caption{\textbf{Performance on T2I-CompBench.}
    Comparison of diffusion-based and visual autoregressive text-to-image
    models across six compositional dimensions.}

    \label{tab:t2icompbench_results}
\end{table*}

\begin{table*}[t]
    \centering
    \scriptsize
    \setlength{\tabcolsep}{3.2pt}
    \renewcommand{\arraystretch}{1.12}

    \begin{tabular}{@{}l|*{6}{c}|*{7}{c}@{}}
        \toprule
    
        \multirow{2}{*}{\textbf{Model}}
        & \multicolumn{6}{c|}{\textbf{Basic}}
        & \multicolumn{7}{c}{\textbf{Advanced}} \\
        \cmidrule(lr){2-7}
        \cmidrule(lr){8-14}
    
        & Attr.
        & Scene
        & Spat.
        & Act.
        & Part
        & Basic
        & Count.
        & Comp.
        & Diff.
        & Neg.
        & Univ.
        & Adv.
        & Overall \\
        \midrule

        \multicolumn{14}{c}{\textbf{Diffusion T2I Models}} \\
        \midrule

        Stable Diffusion v1.4
        & 0.684 & 0.715 & 0.648 & 0.660 & 0.628 & 0.677
        & 0.606 & 0.630 & 0.607 & 0.492 & 0.560 & 0.557 & 0.612 \\

        Stable Diffusion v2
        & 0.723 & 0.766 & 0.697 & 0.688 & 0.663 & 0.717
        & 0.635 & 0.631 & 0.611 & 0.496 & 0.602 & 0.570 & 0.637 \\

        SDXL
        & 0.641 & \textbf{0.855} & 0.820 & \textbf{0.823}
        & 0.831 & \textbf{0.835}
        & 0.725 & 0.690 & 0.709 & 0.511 & 0.667 & 0.641
        & \textbf{0.729} \\

        \midrule
        \multicolumn{14}{c}{
            \textbf{Visual Autoregressive T2I Models}
        } \\
        \midrule

        Infinity-2B
        & 0.856 & 0.867 & 0.860 & 0.850
        & 0.860 & 0.855
        & 0.735 & 0.727 & 0.752 & 0.442 & 0.685 & 0.632
        & 0.734 \\

        Infinity-8B-512
        & 0.885 & 0.893 & 0.885 & 0.873 & 0.882 & 0.884
        & 0.765 & 0.750 & 0.769 & 0.464 & 0.715 & 0.656
        & 0.760 \\

        Infinity-8B
        & \textbf{0.916} & \textbf{0.924} & \textbf{0.917}
        & \textbf{0.908} & \textbf{0.920} & \textbf{0.917}
        & \textbf{0.808} & \textbf{0.794} & \textbf{0.831}
        & \textbf{0.491} & \textbf{0.737} & \textbf{0.695}
        & \textbf{0.796} \\

        \bottomrule
    \end{tabular}

    \caption{\textbf{Performance on GenAI-Bench.}
    Comparison of diffusion-based and visual autoregressive models
    across the basic and advanced evaluation dimensions.
    Attr., Spat., Act., Count., Comp., Diff., Neg., Univ., and Adv.
    denote Attribute, Spatial, Action, Counting, Comparison,
    Differentiation, Negation, Universal, and Advanced, respectively.}
    \label{tab:genaibench_results}
\end{table*}
InfiHall-Bench is complementary to these efforts. We do not introduce a
broader taxonomy of high-level reasoning skills. Instead, we deliberately
retain five foundational skills whose requirements can be stated explicitly
and verified directly from the generated image. This controlled scope reduces
ambiguity about whether a failure originates in language or knowledge
reasoning or in the visual generation process itself. What distinguishes
InfiHall-Bench is its construction procedure: several skills jointly guide
each prompt, and the candidate pool is curated with respect to both
individual-skill coverage and the distribution of skill combinations. The
resulting benchmark tests whether a model can satisfy diverse combinations of
explicit visual constraints, and provides skill-conditioned evidence for
diagnosing which requirement inside a prompt was violated.

\paragraph{Parameter-efficient adaptation.}
CMHM adapts a frozen generator with LoRA~\cite{hu2022lora}, which injects
low-rank updates into selected projections and leaves the pretrained weights
unchanged. Our contribution is orthogonal to the choice of adapter: rather than changing
which parameters are trained, CMHM applies confidence-based weighting within
every autoregressive scale and balances the contributions of different scales,
so that adaptation concentrates on predictions that are not yet reliable.

\section{InfiHall-Bench}
\label{sec:benchmark}
\subsection{Limitations of Existing Benchmarks}

Existing benchmarks present two limitations for diagnosing
hallucinations in increasingly capable VAR models.
Tables~\ref{tab:t2icompbench_results}
and~\ref{tab:genaibench_results} illustrate a shift in their
diagnostic effectiveness. Both benchmarks expose clear weaknesses
of the Stable Diffusion baselines, particularly on spatial and
advanced compositional requirements. However, their scores become
less discriminative for the stronger Infinity configurations.
For example, the Complex score on T2I-CompBench varies only from
0.513 to 0.524 across the three Infinity models, while their Overall
scores on GenAI-Bench range from 0.734 to 0.796. These aggregate
results show that the models remain imperfect, but provide limited
evidence about which explicit requirements fail when several skills
must be satisfied jointly. Similar aggregate performance may therefore
conceal substantially different combinations of attribute, relational,
and scene-level errors.

First, many prompts provide limited challenge and may no longer
expose the remaining weaknesses of these models. Second,
category-driven construction often instantiates prompts through
recurring skill-specific structures, providing limited coverage of
interactions among multiple foundational visual requirements.
Consequently, competitive category-level performance may coexist
with failures when several explicit constraints must be satisfied
jointly. Simply increasing prompt difficulty through implicit
semantics, external knowledge, or higher-order reasoning does not
resolve this issue, because the resulting errors may originate from
prompt understanding or reasoning rather than visual generation.

To provide a challenging yet controlled evaluation, InfiHall-Bench
combines five foundational, explicitly stated, and visually verifiable
skills within coherent scenes. This multi-skill design diversifies
interactions among visual requirements while reducing confounding
from implicit reasoning and external knowledge, enabling more
targeted diagnosis of visual realization failures in VAR models.
\subsection{Prompts of InfiHall-Bench}

The final benchmark contains 411 prompts covering five visual skills. \emph{Attribute binding} evaluates whether visible properties, such as color, material, texture, and state, are assigned to the intended entities. \emph{Spatial relation} evaluates whether explicit layouts, including left/right, above/below, inside/outside, and front/behind, are preserved. \emph{Action relation} evaluates whether the specified subject performs the requested action on the correct object. \emph{Part consistency} evaluates whether a specified object component is present and rendered correctly. \emph{Scene consistency} evaluates whether the objects, properties, relations, and interactions jointly form the requested scene. These skills are not mutually exclusive, and each prompt may be annotated with several of them. In total, the benchmark contains 674 unique entities; each prompt has 4.05 annotated skills on average and an average length of 20.23 words. Scene consistency and action relation each appear in 411 prompts, spatial relation in 337 prompts, attribute binding in 302 prompts, and part consistency in 210 prompts.

\paragraph{Multi-Skill Prompt Generation.}
We use an LLM-assisted prompt constructor to generate candidates around the five skills. Each candidate combines at least two visual skills, contains only content that can be verified from the generated image, and describes a coherent scene using familiar objects and actions. The constructor outputs both the prompt and its associated skill labels. Combining several skills within one prompt increases the likelihood that a model must resolve interacting constraints rather than independently render a single object or attribute.

\paragraph{Quality Control.}
The candidates undergo automatic filtering followed by manual inspection. We first remove malformed prompts and those containing ambiguous, subjective, or non-visual content. Duplicate and near-duplicate prompts are then removed to reduce repeated scenes and linguistic templates. We further inspect the marginal distribution of individual skills and the joint distribution of skill combinations, and rebalance the candidate pool to prevent common combinations from dominating the benchmark. During manual review, a prompt is revised or removed if a relation admits multiple reasonable interpretations, a specified component would not normally be visible, or the assigned skill labels do not cover its main content.

\subsection{Fine-Grained Evaluation Protocol}

For each prompt--image pair, the evaluator receives the generated image, the original prompt, the annotated skill labels, and the corresponding evaluation criteria. We use Qwen3.7-Plus as the multimodal evaluator. It returns an overall score for the complete prompt and a score from 0 to 10 for each annotated skill, together with short rationales for the assigned scores. Skills not annotated for a prompt are not evaluated.

Let $\mathcal{S}$ denote the set of visual skills, and let
$k \in \mathcal{S} \cup \{\mathrm{overall}\}$ denote an evaluation
dimension. We define $\mathcal{P}_{\mathrm{overall}}=\mathcal{P}$,
while $\mathcal{P}_s\subseteq\mathcal{P}$ contains the prompts
annotated with skill $s$. Given the image $I_M(p)$ generated by model
$M$ for prompt $p$, the benchmark score for dimension $k$ is
\begin{equation}
S_k(M)
=
\frac{1}{|\mathcal{P}_k|}
\sum_{p\in\mathcal{P}_k}
E_k\bigl(p,I_M(p)\bigr).
\label{eq:benchmark-score}
\end{equation}
The overall score summarizes prompt--image consistency, while the
per-skill scores identify which visual requirements are not satisfied.
The rationales provide textual descriptions of the corresponding errors.
\section{Method}
\label{sec:method}
\subsection{Motivation}

\begin{figure}[t!]
    \includegraphics[width=\columnwidth]{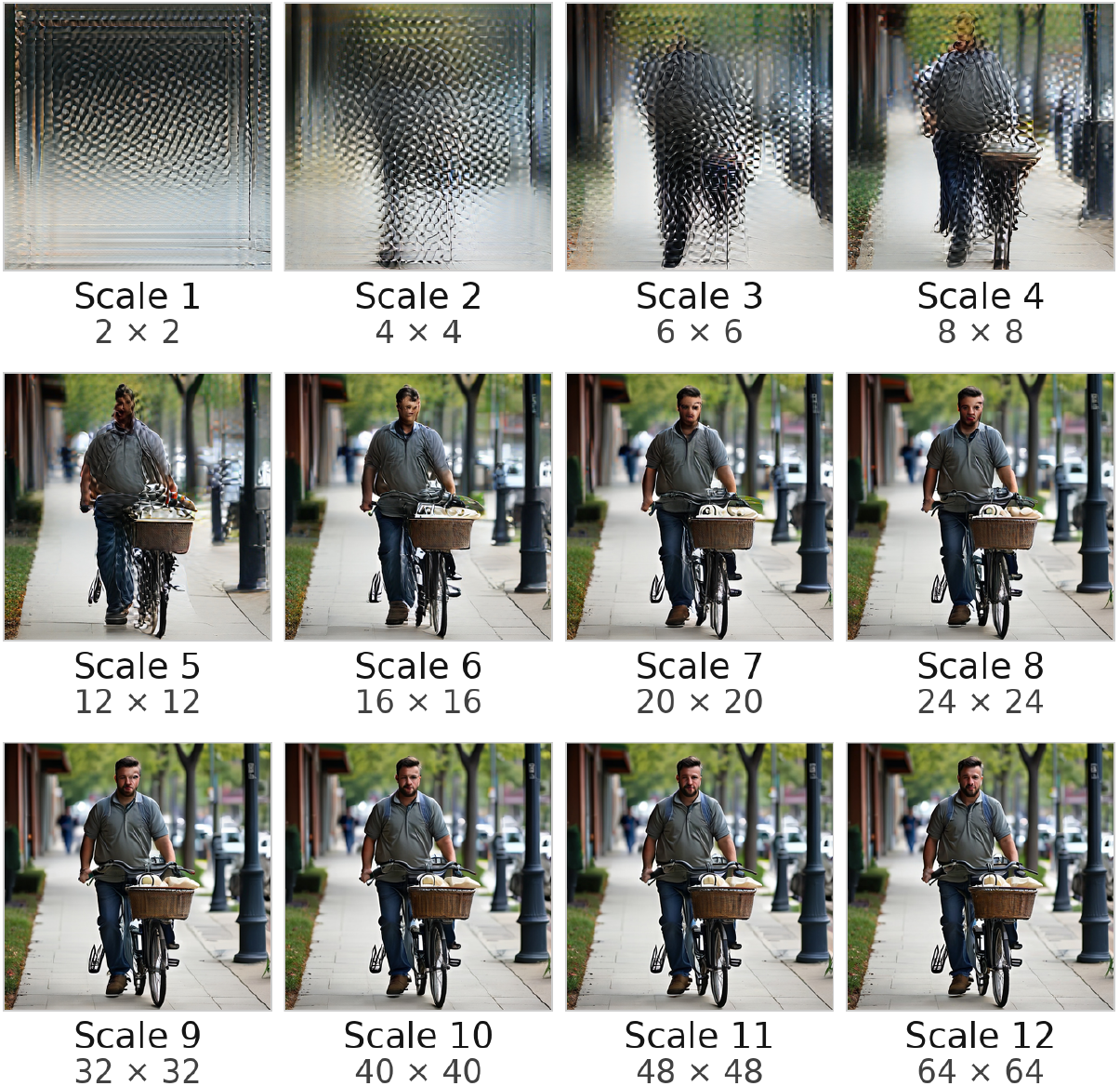}
    \caption{\textbf{Scale-wise persistence of a hallucination in Infinity-2B.} The prompt is: On a sidewalk, a man carries a folded chair behind a bicycle with a basket on the front. By Scale~5 ($12\times12$), the sidewalk scene,
the man, the bicycle, and its front basket have been established, but the
requested action and layout are incorrect: the folded chair is not clearly
realized as an object carried by the man, who instead appears to push or ride
the bicycle. Subsequent scales refine the image without correcting this
semantic error.}
    \label{fig3:method}
\end{figure}

InfiHall-Bench identifies hallucination-prone samples together with the skills
on which they fail. Inspecting their intermediate generations reveals a
characteristic coarse-to-fine failure pattern. As shown in
Fig.~\ref{fig3:method}, once the global composition commits to an
incorrect action or layout, subsequent scales tend to refine that composition
rather than revise it. The scale at which such an error becomes established,
however, varies across samples and requirements. Scale dependence therefore
does not imply that a single predetermined scale should receive all of the
training signal.
Re-encoding the same samples under teacher forcing further shows that the
confidence assigned to target visual tokens is highly nonuniform both within
and across scales. Standard fine-tuning weights all visual tokens uniformly,
allowing the many tokens that the model already predicts reliably to dominate
the objective. We therefore propose Confidence-Masked Hallucination Mitigation
(CMHM), which applies confidence-based weighting at every autoregressive scale
and emphasizes uncertain tokens wherever they occur. Per-scale normalization
prevents fine scales, which contain more tokens, from dominating the update.
CMHM does not assume that low confidence exactly identifies a hallucinated
token; instead, confidence serves as an inexpensive proxy for allocating
limited adapter capacity to predictions that the model has not yet learned
reliably and that are therefore more likely to encode unsatisfied prompt
requirements.

\subsection{Bitwise Autoregressive Fine-Tuning}
Given a prompt--image pair $(p,x)$, the Infinity visual tokenizer represents
$x$ as a coarse-to-fine sequence of bitwise tokens $\{y_t\}_{t=1}^{T}$, where
$y_t\in\{0,1\}^{D}$ and $D$ is the number of bits per token. Let
$s(t)\in\{1,\ldots,S\}$ denote the visual scale containing token $t$. We
insert LoRA adapters~\cite{hu2022lora} into the autoregressive transformer and
optimize only the adapter parameters, leaving the pretrained parameters
$\theta_0$ frozen. Under teacher forcing, the bit-averaged prediction loss for
token $t$ is
\begin{equation}
    \ell_t=\frac{1}{D}\sum_{j=1}^{D}
    \mathrm{CE}\bigl(z_{t,j},y_{t,j}\bigr),
    \label{eq:token_loss}
\end{equation}
where $z_{t,j}$ is the predicted logit for the $j$-th bit conditioned on the
prompt and the preceding ground-truth tokens.

\subsection{Confidence-Based Token Masking}

For each target bit, we obtain the probability assigned to its ground-truth
value. The confidence of token $t$ is the mean target-bit probability
\begin{equation}
    c_t=\frac{1}{D}\sum_{j=1}^{D}
    P_{\theta}\bigl(y_{t,j}\mid y_{<t},p\bigr).
    \label{eq:confidence}
\end{equation}
We convert $c_t$ into a soft loss weight using two thresholds,
$\tau_{\mathrm{low}}<\tau_{\mathrm{high}}$:
\begin{equation}
    m_t=\operatorname{clip}\left(
    \frac{\tau_{\mathrm{high}}-c_t}
    {\tau_{\mathrm{high}}-\tau_{\mathrm{low}}},0,1\right).
    \label{eq:mask}
\end{equation}
Tokens with $c_t\leq\tau_{\mathrm{low}}$ keep their full loss, tokens with
$c_t\geq\tau_{\mathrm{high}}$ are masked out, and tokens between the two
thresholds are weighted linearly. Compared with a hard selection rule, the
soft transition avoids an abrupt change in the training signal near either
threshold. We refer to the tokens with $m_t>0$ as the \emph{retained} tokens.

To prevent scales that contain more tokens from dominating the objective, we
aggregate the masked loss within each scale and normalize it by the total
weight of the retained tokens in that scale:
\begin{equation}
    \bar{\mathcal{L}}_s=
    \frac{\sum_{t:s(t)=s}m_t\ell_t}
    {\sum_{t:s(t)=s}m_t+\epsilon},
    \qquad
    \mathcal{L}_{\mathrm{CMHM}}=
    \sum_{s=1}^{S}\omega_s\bar{\mathcal{L}}_s,
    \label{eq:CMHM_loss}
\end{equation}
where $\omega_s$ controls the contribution of scale $s$ and $\epsilon$ ensures
numerical stability. The denominator keeps the average loss and gradient scale
comparable to uniform training, so masking changes only the relative
contribution of individual tokens; confidence selects uncertain predictions
within a scale, while $\omega_s$ balances the scales against each other.

To limit distributional drift, we additionally regularize the fine-tuned model
on clean replay samples~\cite{hinton2015distilling}, that is, prompt--image pairs on which the pretrained
model already satisfies every annotated requirement. The regularizer is the
token-level divergence
\begin{equation}
    \mathcal{L}_{\mathrm{reg}}=
    \frac{1}{T}\sum_{t=1}^{T}
    \mathrm{KL}\!\left(
    P_{\theta_0}(\cdot\mid y_{<t},p)
    \;\|\;
    P_{\theta}(\cdot\mid y_{<t},p)
    \right).
    \label{eq:replay}
\end{equation}
The final training objective is
\begin{equation}
    \mathcal{L}=\mathcal{L}_{\mathrm{CMHM}}+
    \lambda_{\mathrm{reg}}\mathcal{L}_{\mathrm{reg}}.
    \label{eq:full_objective}
\end{equation}
CMHM modifies only the allocation of the training loss across visual tokens
and scales. It requires no architectural change and adds no module at
inference time.
\section{Experiments}
\begin{table*}[t!]
    \centering
    \setlength{\tabcolsep}{9.0pt}
    \begin{tabular}{l|cccccc}
        \toprule
        \textbf{Model}
        & Action Rel.
        & Attribute
        & Part
        & Scene
        & Spatial Rel.
        & Overall \\
        \midrule  
        \multicolumn{7}{c}{\textbf{Diffusion T2I Models}} \\
        \midrule
        SDXL
        & 40.80 & 68.91 & 55.40 & 87.66 & 33.03 & 47.49 \\
        PixArt-Sigma-XL-2-1024-MS 
        & 45.74 & 76.36 & 54.43 & 85.44 & 39.94 & 51.76 \\
        SD3.5-Medium 
        & 58.28 & 85.76 & 67.40 & 91.02 & 52.46 & 61.68 \\
        SD3.5-Large 
        & 61.90 & 86.57 & 75.62 & \textbf{93.77} & 53.32 & 64.04 \\
        FLUX.1-dev 
        & \textbf{66.85} & \textbf{90.35} & \textbf{75.70}
        & 92.87 & \textbf{54.13} & \textbf{67.08} \\
        
        \midrule
        \multicolumn{7}{c}{\textbf{Autoregressive T2I}} \\
        \midrule
        
        Janus-Pro-7B
        & 37.28 & 66.43 & 55.15 & 78.25 & 43.29 & 46.69 \\
        LlamaGen 
        & 32.38 & 45.80 & 33.45 & 64.57 & 23.26 & 34.26 \\
        
        \midrule
        \multicolumn{7}{c}{\textbf{Finetuned Visual Autoregressive T2I}} \\
        \midrule
        
        Infinity-2B
        & 48.40 & 77.77 & 54.81 & 87.64 & 39.02 & 50.83 \\
        \quad + infihall-train
        & 50.28 & \textbf{80.21} & \textbf{63.96}
        & 93.46 & 45.64 & 57.30 \\
        \quad + CMHM (Ours)
        & \textbf{53.05} & 80.11 & 63.75
        & \textbf{93.65} & \textbf{47.77} & \textbf{58.13} \\
        
        \midrule
        
        Infinity-8B-512
        & 53.05 & 82.12 & 62.55 & 90.90 & 42.70 & 56.42 \\
        \quad + infihall-train
        & 60.53 & \textbf{84.98} & 69.66
        & \textbf{95.18} & 53.77 & 63.80 \\
        \quad + CMHM (Ours)
        & \textbf{61.08} & 83.85 & \textbf{70.00}
        & 95.06 & \textbf{56.38} & \textbf{64.77} \\
        
        \midrule
        
        Infinity-8B
        & 64.53 & 88.94 & \textbf{76.55}
        & \textbf{96.08} & 56.56 & 65.77 \\
        \quad + infihall-train
        & 63.88 & 86.50 & 70.12
        & 95.21 & 58.99 & 66.35 \\
        \quad + CMHM (Ours)
        & \textbf{66.28} & \textbf{89.05} & 72.04
        & 95.94 & \textbf{60.45} & \textbf{67.49} \\       
        \bottomrule 
    \end{tabular}
    \caption{\textbf{Results on InfiHall-Bench.} Comparison of diffusion-based and visual autoregressive T2I models using the Overall and five skill-specific scores. For each Infinity model, we report the pretrained, infihall-train, and CMHM results.}
    \label{tab:main}
\end{table*}
\subsection{Experimental Setup}
\paragraph{Evaluated models.}We evaluate five diffusion-based models, including SDXL~\cite{podell2024sdxl}, PixArt-$\Sigma$~\cite{chen2024pixart}, Stable Diffusion 3.5 Medium~\cite{esser2024scaling} and Large~\cite{esser2024scaling}, and FLUX.1-dev~\cite{flux2024}, together with Janus-Pro-7B~\cite{chen2025janus} and three Infinity configurations on the 411 prompts of InfiHall-Bench. 
\paragraph{Evaluation Details}For each generated image, Qwen3.7-Plus evaluates overall prompt fidelity and the annotated skills, including action relation, attribute, part, scene, and spatial relation. All scores range from 0 to 10. The Overall score is averaged over the complete benchmark, whereas each skill score is computed only from prompts annotated with that skill.
\paragraph{Training Details}For CMHM, we set $\tau_{\mathrm{low}}=0.3$, $\tau_{\mathrm{high}}=0.7$, and $\beta=2.0$, and apply confidence-based weighting at all autoregressive scales. The resulting weights are normalized to have unit mean over the retained tokens. Each training step also includes one clean replay example, with the KL regularization weight set to $\lambda_{\mathrm{KL}}=0.01$. For a controlled comparison, the infihall-train baseline uses the same training data, adapter configuration, optimization settings, and replay regularization as CMHM, but applies uniform token weighting. All fine-tuning experiments are conducted on two NVIDIA RTX PRO 6000 Blackwell GPUs with 96\,GB of memory each, with each training run using a single GPU.

\subsection{Hallucination Diagnosis}
Table~\ref{tab:main} reports the results on InfiHall-Bench. Among the pretrained models, FLUX.1-dev achieves the highest Overall score of 67.08, while Infinity-8B reaches a comparable score of 65.77 and substantially outperforms the other VAR models. Across the three pretrained Infinity configurations, the Overall score increases from 50.83 to 65.77, showing that stronger configurations improve prompt fidelity but do not eliminate fine-grained hallucinations. In particular, even Infinity-8B obtains only 64.53 and 56.56 on Action Relation and Spatial Relation, respectively, despite achieving 96.08 on Scene Consistency. This large gap indicates that generating a plausible global scene is considerably easier than faithfully realizing the actions and relative layouts specified in the prompt.

CMHM achieves Overall scores of 58.13, 64.77, and 67.49 on Infinity-2B, Infinity-8B-512, and Infinity-8B, improving their pretrained counterparts by 7.30, 8.35, and 1.72 points, respectively. The larger gains on Infinity-2B and Infinity-8B-512 suggest that hallucination-oriented adaptation provides greater absolute benefits when the pretrained configuration has more remaining errors. Nevertheless, CMHM also improves the strongest Infinity-8B configuration and raises its Overall score above FLUX.1-dev by 0.41 points. A closer comparison reveals different skill profiles: Infinity-8B with CMHM exceeds FLUX.1-dev by 6.32 points on Spatial Relation and 3.07 points on Scene Consistency, while remaining slightly lower on Action Relation, Attribute, and Part Consistency. Thus, similar Overall scores can conceal substantially different hallucination patterns, further motivating the skill-level evaluation provided by InfiHall-Bench.

More importantly, CMHM consistently outperforms the corresponding infihall-train baseline, improving the Overall score by 0.83, 0.97, and 1.14 points across the three configurations. Specifically, it improves Action Relation by 2.77, 0.55, and 2.40 points and Spatial Relation by 2.13, 2.61, and 1.46 points on Infinity-2B, Infinity-8B-512, and Infinity-8B, respectively. These consistent relational gains support the motivation of CMHM: uncertain visual tokens are more likely to encode difficult interactions and layouts that uniform fine-tuning does not emphasize sufficiently. Improvements in the remaining skills are model dependent, indicating that confidence-based weighting primarily reallocates optimization toward challenging relational requirements rather than uniformly improving every visual property.

\subsection{Further Analysis}
Additional qualitative comparisons are provided in the appendix, covering the pretrained, infihall-train, and CMHM variants of all three Infinity backbones. These examples complement the quantitative results by showing how CMHM improves the realization of specific prompt requirements. The appendix also includes further details on benchmark construction, evaluation, and training-data generation. The appendix also presents additional ablation studies to examine the contribution of the key components in CMHM.
\section{Conclusion}
We present InfiHall-Bench, a benchmark for evaluating hallucinations in visual autoregressive text-to-image generation. The benchmark comprises 411 multi-skill prompts spanning attribute binding, spatial relations, action relations, part consistency, and scene consistency. Our experiments reveal that Infinity models remain competitive on existing compositional benchmarks yet continue to violate a substantial number of fine-grained visual requirements captured by InfiHall-Bench, particularly those involving spatial and action relations. To mitigate these hallucinations, we further introduce Confidence-Masked Hallucination Mitigation, which combines LoRA adaptation with confidence-based token masking to focus optimization on uncertain visual tokens, while employing clean-replay regularization to mitigate distributional drift. InfiHall-Bench establishes a systematic framework for evaluating hallucinations in visual autoregressive generation.
\bigskip
\bibliography{aaai2027}
\onecolumn
\section{Appendix}
\label{appendix}

\subsection{Benchmark Construction Details}
To construct the evaluation prompts for InfiHall-Bench, we employ an LLM-assisted generation pipeline. 
The goal is to create challenging yet visually verifiable prompts that can expose text-image alignment failures in advanced text-to-image models, while avoiding ambiguous or unrealistic scenarios.

We provide the LLM with the following instruction:

\textbf{System Prompt}:

\begin{lstlisting}[breaklines=true, basicstyle=\small\ttfamily, frame=single, numbers=none]
You are generating hard basic evaluation prompts for a text-to-image benchmark.
Your task is to create diverse, visually grounded, and image-verifiable prompts for evaluating text-image alignment in strong text-to-image models.
These prompts will be used for hallucination analysis and mitigation research on basic visual consistency errors.
Goal:
Create prompts that are likely to expose failures in Infinity-2B, Infinity-8B-512, and Infinity-8B,
while still using common objects, familiar scenes, and visually clear constraints.
The prompts should be hard but fair: do not rely on rare concepts, obscure knowledge, unnatural situations, or overly complex reasoning.
Requirements:
1. Each prompt must include two to four basic skill types from the allowed skill list.
2. The prompt should be visually clear, natural, and verifiable from the image alone.
3. The prompt should be challenging but not ambiguous, contradictory, or impossible to visualize.
4. Use common and recognizable objects, people, animals, scenes, and actions.
5. Avoid rare objects, obscure cultural references, brand names, specialized tools, or expert-only concepts.
6. Avoid abstract concepts, subjective judgments, hidden properties, or non-visual knowledge.
7. Prefer one sentence per prompt, ideally 10 to 25 words.
8. Each prompt should contain two to four clear visual constraints, such as an attribute, action, scene, spatial relation, or visible part.
9. Do not overload a prompt with too many objects or relations.
10. Avoid unnecessary style, camera, lighting, or artistic details unless required by the listed skills.
11. Use diverse objects, scenes, attributes, actions, parts, and wording. Avoid repeated templates and near-duplicates.
12. Do not copy or lightly paraphrase existing benchmark prompts.
13. Do not use object as a skill label; objects should naturally appear in the prompt text.
14. Avoid advanced skills such as negation, counting, comparison, text rendering, symbolic reasoning, or complex logic.
15. Each prompt must have a clear correct visual interpretation.
Allowed basic skill types:
attribute, scene, spatial_relation, action_relation, part.
Skill definitions:
attribute = visible properties such as color, shape, size, texture, material, or pattern bound to the correct object;
scene = a clear visual environment or location;
spatial_relation = visible positional relations such as left/right, above/below, in front of/behind, inside/outside, under, beside, or between;
action_relation = a subject performing a visible action;
part = visible object parts or components.
\end{lstlisting}
During generation, we additionally provide a lightweight user template to control batch generation and maintain diversity across rounds. 
The template specifies the number of prompts to generate, the current generation round, and previously generated prompts for duplicate filtering.

\subsection{Evaluation Details}
We use Qwen-3.7-plus to evaluate generation results. It uses the following context as instruction:
\begin{lstlisting}[breaklines=true, basicstyle=\small\ttfamily, frame=single, numbers=none]
You are a STRICT image-text hallucination evaluator for text-to-image models.
Your task is to evaluate whether the image faithfully follows the given prompt,
with a focus on the specified visual skills.
You MUST follow the scoring format exactly.
------------------------------------------------------------
## SCORING RULES
------------------------------------------------------------
### 1. Overall Score
Overall score reflects global image-prompt consistency:
- 10: perfect match, no hallucination
- 7-9: minor errors, mostly correct
- 4-6: partial mismatch or noticeable missing elements
- 1-3: strong hallucinations or missing key content
- 0: completely irrelevant image
""" + (PER_SKILL_SCORING_RULES if EVALUATE_PER_SKILL else OVERALL_ONLY_SCORING_RULES) + """
------------------------------------------------------------
## EVALUATION PRINCIPLES
------------------------------------------------------------
- Only use visual evidence from the image.
- Do NOT assume missing details.
- Do NOT use external knowledge.
- Do NOT interpret beyond what is clearly visible.
- Be conservative: when uncertain, lower the score.
------------------------------------------------------------
## IMPORTANT CONSTRAINTS
------------------------------------------------------------
""" + (PER_SKILL_CONSTRAINTS if EVALUATE_PER_SKILL else "") + """
- Keep reasons short and visual-focused.
- Avoid speculation or storytelling.
\end{lstlisting}
\subsection{Train Details}
To construct the training prompts for hallucination mitigation, we employ an LLM-assisted generation process to create visually grounded and challenging text-to-image prompts. 
Different from the benchmark prompt construction, these prompts are designed for LoRA fine-tuning and focus on basic text-image consistency errors.

The LLM is provided with the following instruction:

\textbf{Training Prompt Generation}:

\begin{lstlisting}[breaklines=true, basicstyle=\small\ttfamily, frame=single, numbers=none]
You are generating basic text-to-image training prompts for hallucination mitigation.
Goal:
create visually clear and moderately challenging prompts for LoRA fine-tuning of a text-to-image model,
focusing on basic text-image consistency errors.

Requirements:
1. Generate new prompts that focus on basic hallucination types, including attributes, scenes, spatial relations, actions, and object parts.
2. Each prompt must be visually grounded, image-generative, and easy to verify from the generated image.
3. Avoid abstract concepts, subjective descriptions, or knowledge that cannot be clearly checked visually.
4. Each prompt should include one to four basic skill types.
5. Do not distinguish between primary and secondary skills; all listed skills should be meaningfully reflected in the prompt.
6. The skill types should be diverse across the batch; do not make all prompts focus on the same skill combination.
7. Each prompt should contain a small number of clear visual constraints; do not overload one prompt with too many constraints.
8. Make the prompt challenging but not contradictory.
9. Prefer one sentence, 8-20 words.
10. Do not include camera style, artistic style, or unnecessary background details unless they are relevant to the skill.
11. The generated prompt should have a clear correct visual answer.
12. Return results in JSONL format only.

Skill types:
attribute, scene, spatial_relation, action_relation, part.

Skill definitions:
attribute = visible properties such as color, shape, size, texture, material, or pattern bound to the correct object;
scene = a clear visual environment or location;
spatial_relation = visible positional relations such as left/right, above/below, in front of/behind, inside/outside, or between;
action_relation = a subject performing a clear visual action;
part = visible object parts or components.
\end{lstlisting}

During generation, we additionally provide the number of prompts to generate, the current generation round, and previously generated prompts for duplicate filtering.

\subsection{Additional Experimental Results}
\begin{figure}
    \centering
    \includegraphics[width=0.9\textwidth]{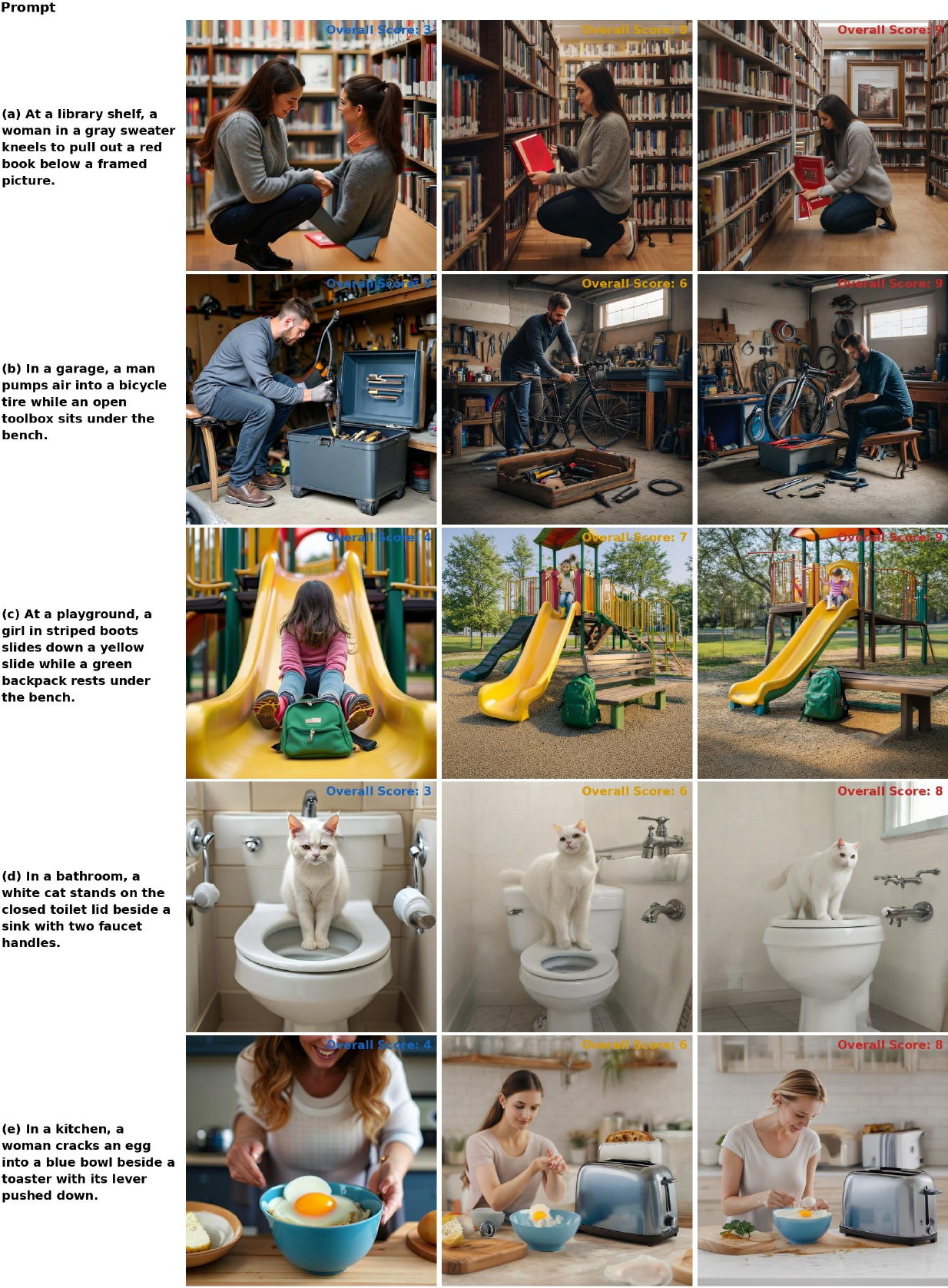}
    \caption{Result of Infinity-2b. The images from left to right belong to baseline,lora,CMHM,which show that the scores are getting higher.}
    \label{fig:2b}
\end{figure}

\begin{figure}
    \centering
    \includegraphics[width=0.9\textwidth]{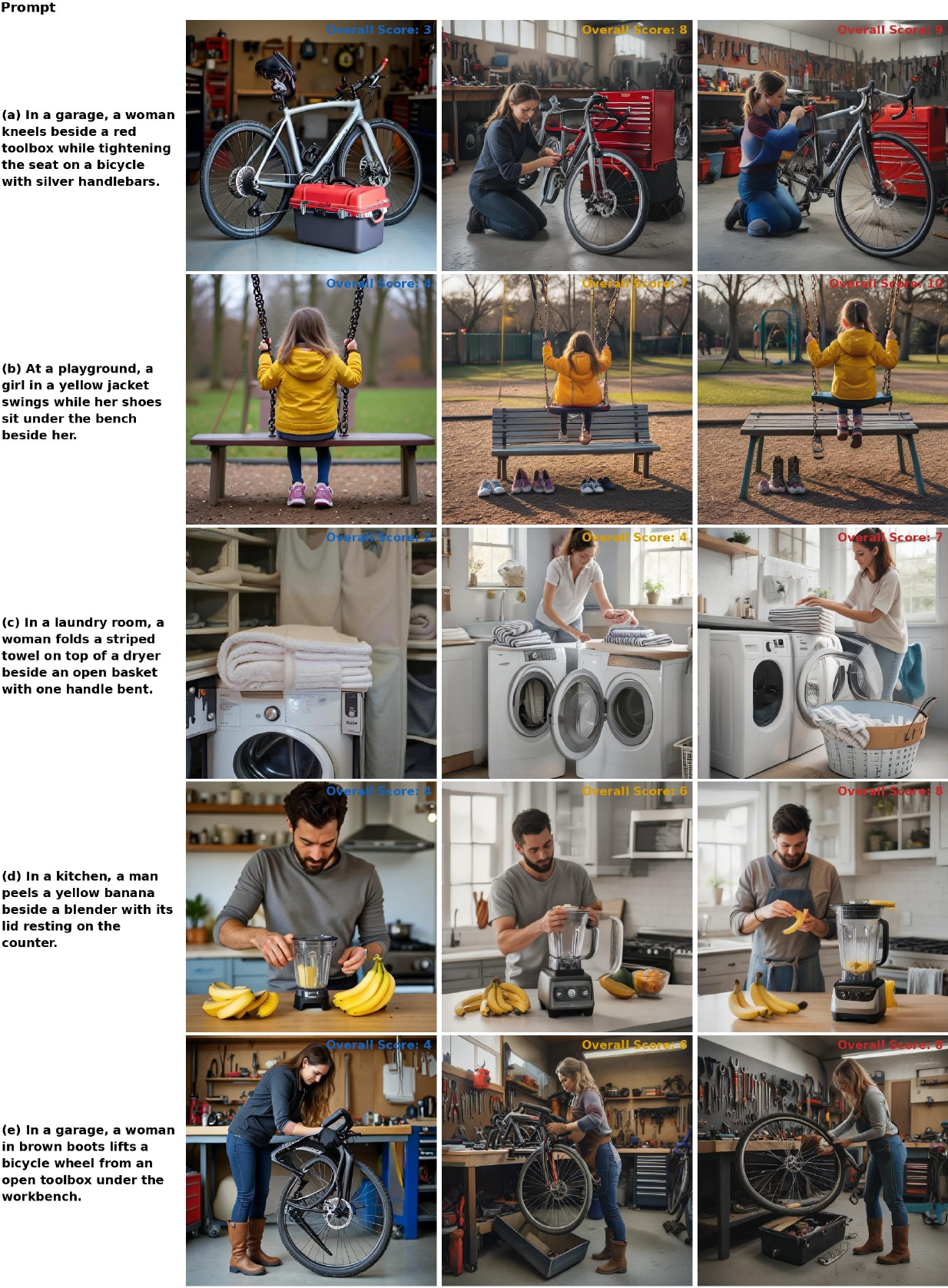}
    \caption{Result of Infinity-8b-512. The images from left to right belong to baseline,lora,CMHM,which show that the scores are getting higher.}
    \label{fig:8b512}
\end{figure}

\begin{figure}
    \centering
    \includegraphics[width=0.9\textwidth]{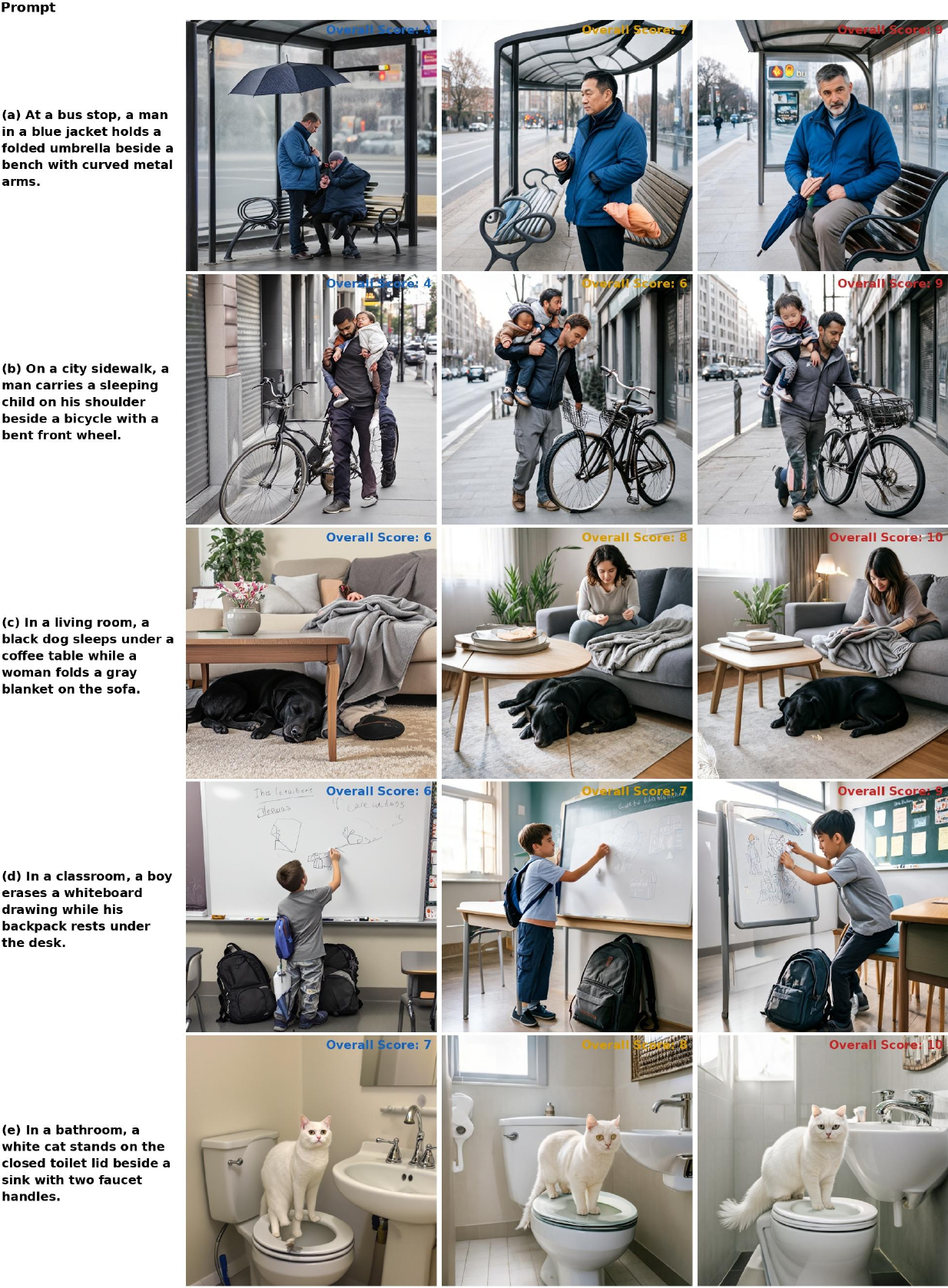}
    \caption{Result of Infinity-8b. The images from left to right belong to baseline,lora,CMHM,which show that the scores are getting higher.}
    \label{fig:8b1m}
\end{figure}

\begin{figure}
    \centering
    \includegraphics[width=1.0\linewidth]{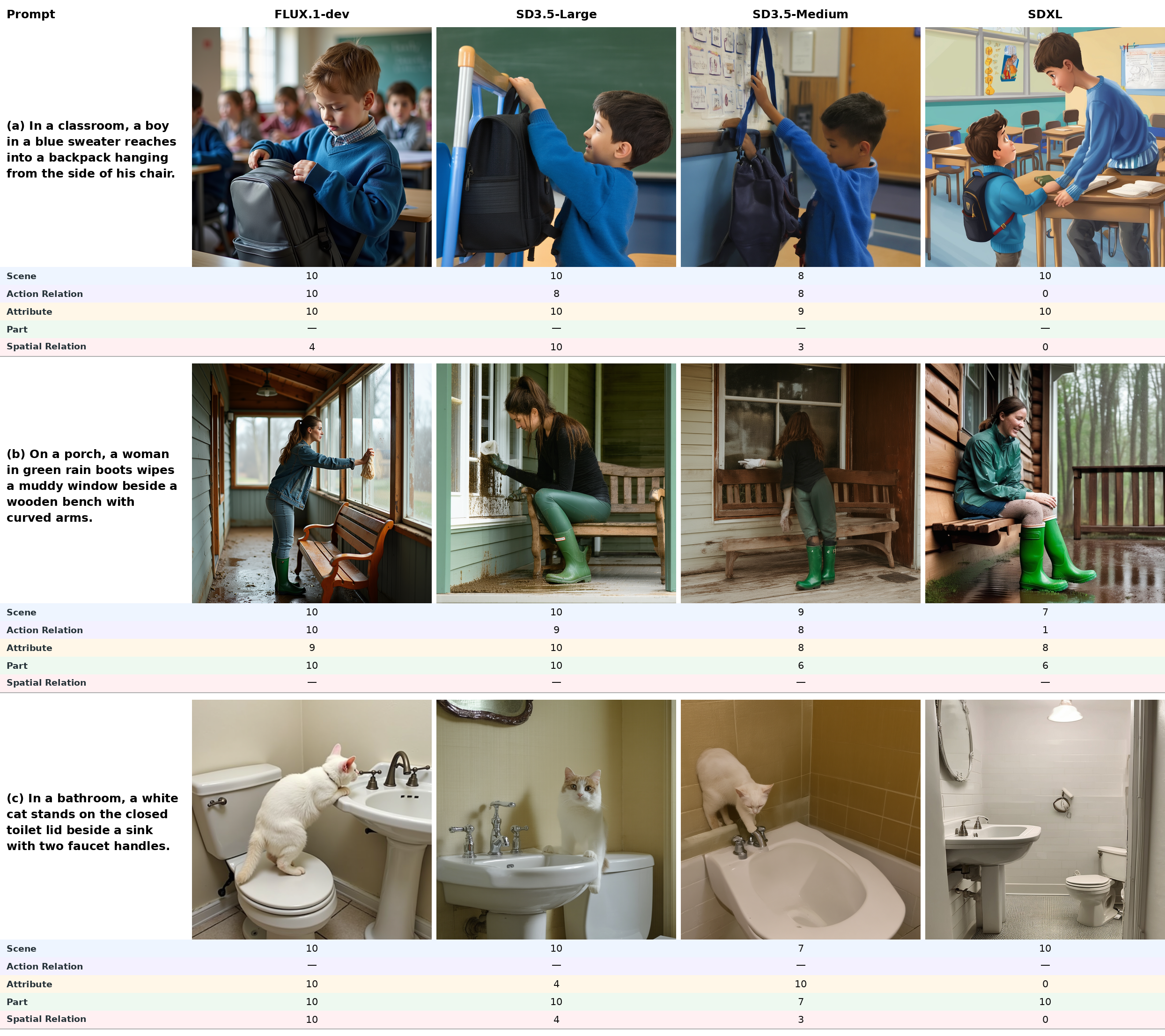}
    \caption{Results of Diffusion T2I Models in Table 3. These images show different diffusions' performance.}
    \label{fig:placeholder}
\end{figure}

\begin{figure}
    \centering
    \includegraphics[width=0.9\linewidth]{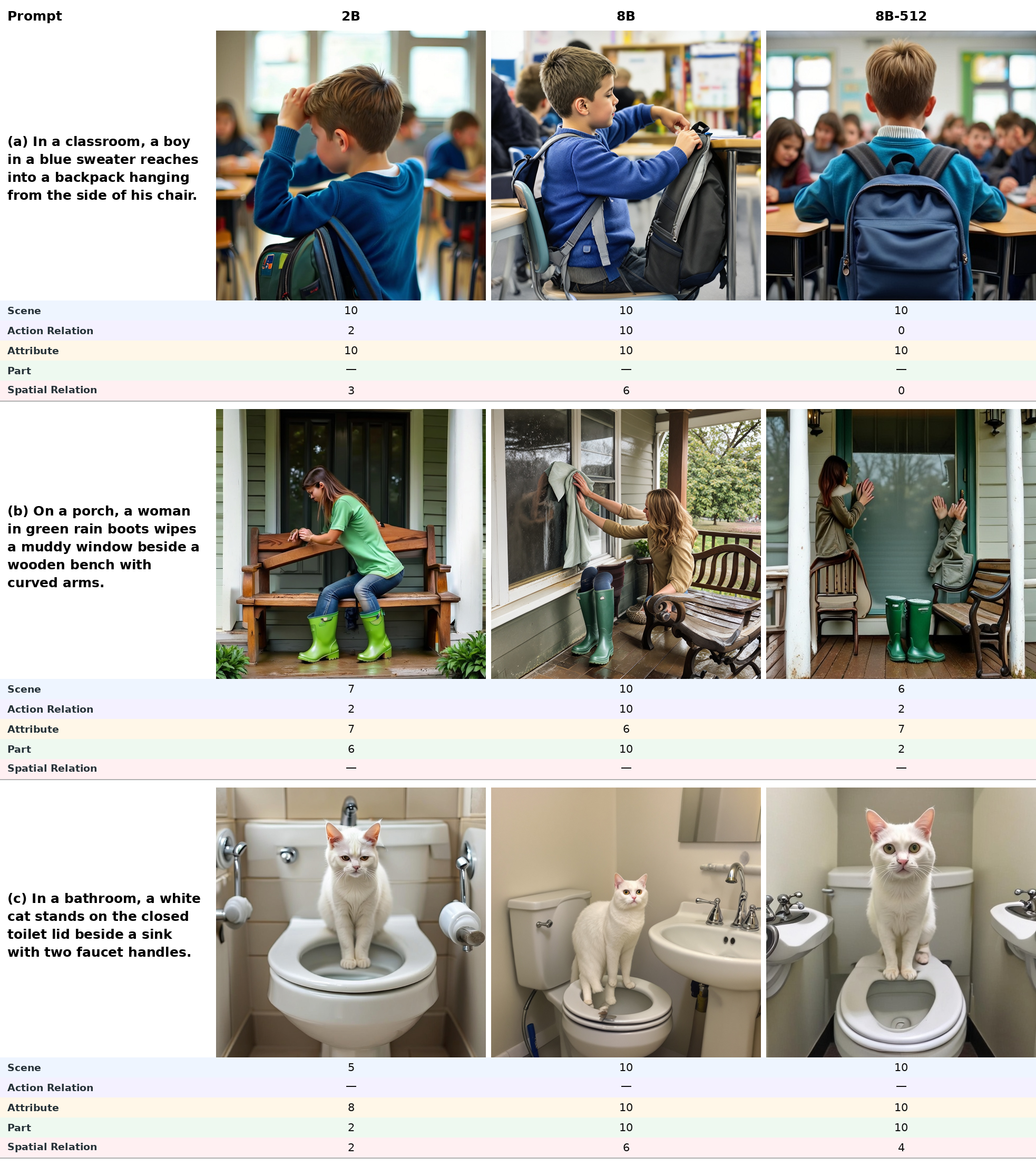}
    \caption{Results of Infinity baseline model in Table 3.These images reveal difference between Infinity models.Infinity-8b is a better one among of them.}
    \label{fig:placeholder}
\end{figure}

\begin{figure}
    \centering
    \includegraphics[width=0.9\linewidth]{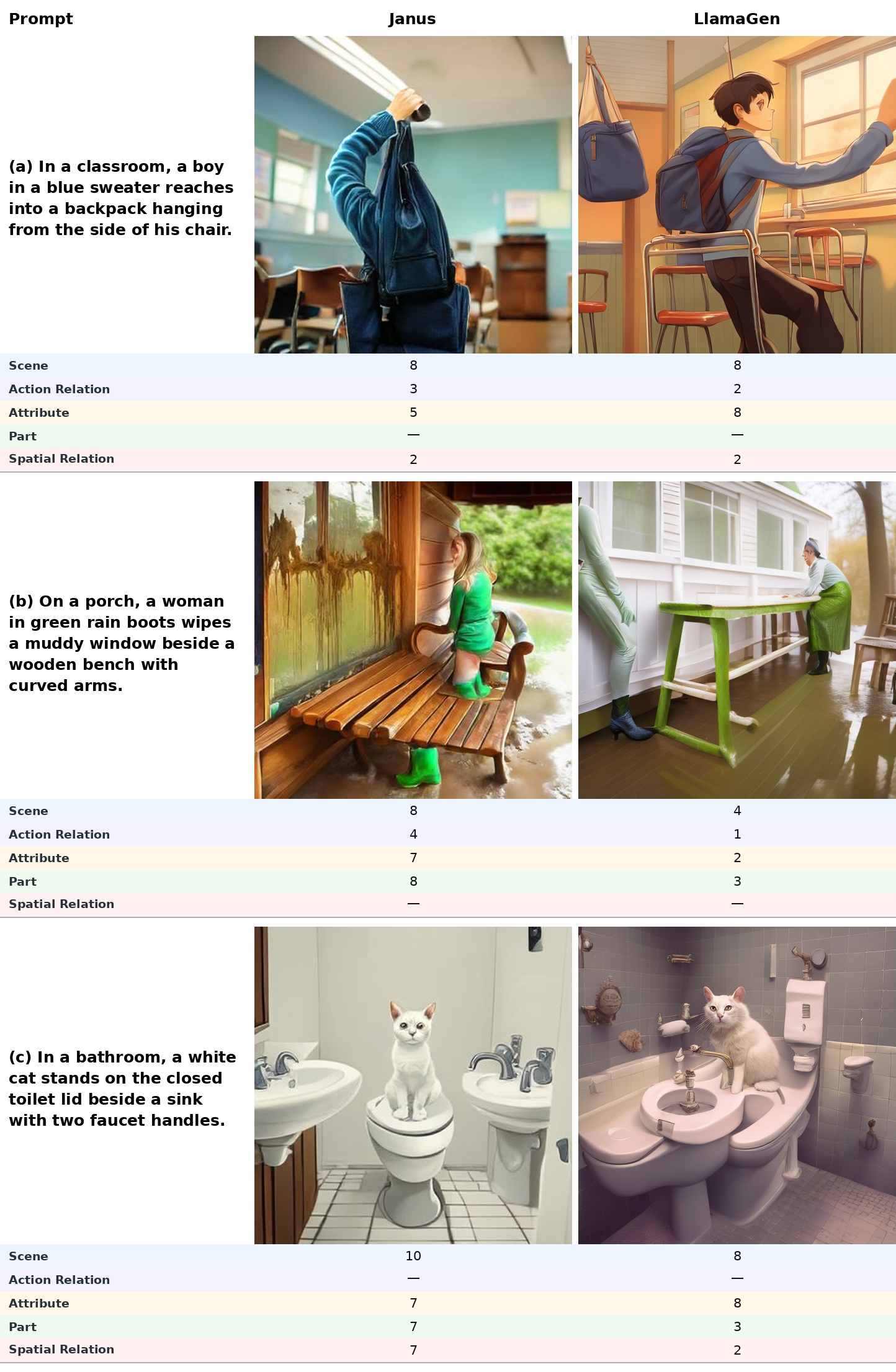}
    \caption{Results of AR T2I Models in Table 3. These images show different AR models' performance.}
    \label{fig:placeholder}
\end{figure}

\end{document}